\documentclass{article}

\PassOptionsToPackage{numbers, compress}{natbib}
 \usepackage[preprint]{neurips_2026}

\usepackage[utf8]{inputenc} 
\usepackage[T1]{fontenc}    
\usepackage{hyperref}       
\usepackage{url}            
\usepackage{booktabs}       
\usepackage{amsfonts}       
\usepackage{nicefrac}       
\usepackage{microtype}      
\usepackage{xcolor}         

\usepackage{amsmath}
\usepackage{graphicx} 
\usepackage{multirow}

\title{Exploring SAM Supervision for Fine-Grained UAV Target Segmentation under Data Scarcity}

\author{%
  Le-Anh Tran \\
  \texttt{tranlevision@gmail.com} \\
}

\begin{document}

\maketitle

\begin{abstract}
Unmanned aerial vehicle (UAV) target segmentation remains challenging due to the small size of objects, large appearance variations, cluttered backgrounds, and the scarcity of densely annotated training data. These factors hinder the performance and practical deployment of lightweight segmentation models in real-world UAV applications. To address this problem, this paper investigates the use of \textbf{SAM3} (Segment Anything Model 3) as a pseudo-label generator for training compact segmentation networks. Specifically, two supervision paradigms are explored: (i) direct pseudo-supervision using unaltered SAM3-generated masks, and (ii) a refinement strategy that re-applies SAM3 to localized image patches for improved mask quality. Based on these paradigms, a two-stage SAM3-guided pseudo-label generation framework is proposed. In the first stage, SAM3 generates coarse masks for initial object localization. The localized regions are subsequently cropped into image patches and processed by SAM3 again to generate fine masks with more accurate object boundaries and discard false positives. The resulting coarse and fine masks are then used as pseudo-labels to optimize a lightweight segmentation network, termed \textbf{IPS-Seg}, which consists of three complementary components: an \textbf{Identity}Former backbone for efficient feature extraction, an Atrous Spatial Pyramid \textbf{Pooling} module for multi-scale context aggregation, and a Pixel\textbf{Shuffle}-based decoder for progressive spatial resolution recovery. Extensive experiments under multiple supervision settings demonstrate the effectiveness of the proposed framework. The results show that IPS-Seg achieves a favorable trade-off between segmentation accuracy and computational efficiency while benefiting consistently from the proposed pseudo-label generation strategy. These findings highlight the potential of leveraging large-scale foundation models as annotation sources for training compact task-specific segmentation networks in low-label vision domains.
\end{abstract}

\section{Introduction}
\label{sec:intro}

Semantic segmentation is a fundamental task in computer vision and plays an important role in unmanned aerial vehicle (UAV) perception systems \cite{kim2025semantic,do2025improved}. In particular, accurate segmentation of UAVs captured by onboard drone cameras is essential for applications such as autonomous navigation, collision avoidance, aerial surveillance, and airspace monitoring. However, unlike conventional aerial-scene segmentation, UAV-target segmentation remains highly challenging because the target drones are typically small, fast-moving, and captured under dynamic imaging conditions with complex backgrounds, illumination variations, and occlusions \cite{kim2025semantic}. Moreover, constructing large-scale pixel-level annotated datasets is expensive due to significant appearance variations across viewpoints, scales, distances, and environmental conditions. As a result, conventional fully supervised segmentation approaches often struggle to achieve robust generalization under limited-label scenarios.

Recent breakthroughs in large-scale vision foundation models have provided promising opportunities for alleviating annotation scarcity in dense prediction tasks. In particular, the Segment Anything Model (SAM) family~\cite{kirillov2023segment, ravi2025sam, carion2025sam} has demonstrated strong zero-shot and prompt-driven segmentation capabilities across diverse visual domains. Among them, the recently introduced SAM3~\cite{carion2025sam} further extends segmentation performance through open-vocabulary semantic understanding and flexible prompt mechanisms. Such capabilities enable robust object localization and mask generation without requiring task-specific retraining. The primary challenge lies in the impracticality of directly deploying SAM-based architectures on resource-constrained UAV platforms due to their high computational complexity. However, SAM3 can serve as a promising supervision source under low-label conditions.

This paper proposes a SAM3-guided pseudo-labeling framework for lightweight UAV-target segmentation under annotation-scarce scenarios. The framework is investigated under two configurations: a one-stage strategy for coarse segmentation and a two-stage strategy designed to exploit the complementary strengths of semantic localization and fine-grained boundary refinement. In the one-stage setting, SAM3 is directly prompted to generate segmentation masks, which are subsequently used as pseudo-labels for training a lightweight segmentation network. In the two-stage setting, the initial SAM3 predictions are first utilized to localize UAV targets, then the obtained localization cues are used to extract informative image patches, which are processed again by SAM3 to produce finer segmentation masks with improved boundary. In addition, a lightweight segmentation network, termed \textbf{IPS-Seg}, is proposed. The proposed architecture consists of three key components: an \textbf{Identity}Former backbone \cite{yu2023metaformer} for efficient feature extraction, an Atrous Spatial Pyramid \textbf{Pooling} (ASPP) bottleneck \cite{chen2017deeplab} for multi-scale contextual aggregation, and a Pixel\textbf{Shuffle}-based decoder \cite{shi2016real} for progressive spatial resolution recovery. Skip connections from the encoding stages are further incorporated to preserve fine-grained boundary information during decoding. Owing to its lightweight yet effective design, IPS-Seg achieves a favorable balance between segmentation accuracy and computational efficiency. Experiments under the fully supervised setting on the UAV Semantic Segmentation dataset~\cite{kim2025semantic} demonstrate that IPS-Seg is well suited for resource-constrained applications. As illustrated in Fig.~\ref{fig:tradeoff}, IPS-Seg delivers competitive segmentation performance while maintaining a compact model size and low computational cost. Furthermore, experiments under pseudo-label supervision demonstrate the effectiveness of the proposed SAM3-guided supervision framework in data-scarce scenarios, highlighting the potential of foundation model supervision for efficient UAV target segmentation under limited annotation budgets. The main contributions of this work are summarized as follows:

\begin{figure}
\centering
\includegraphics[width=\textwidth]{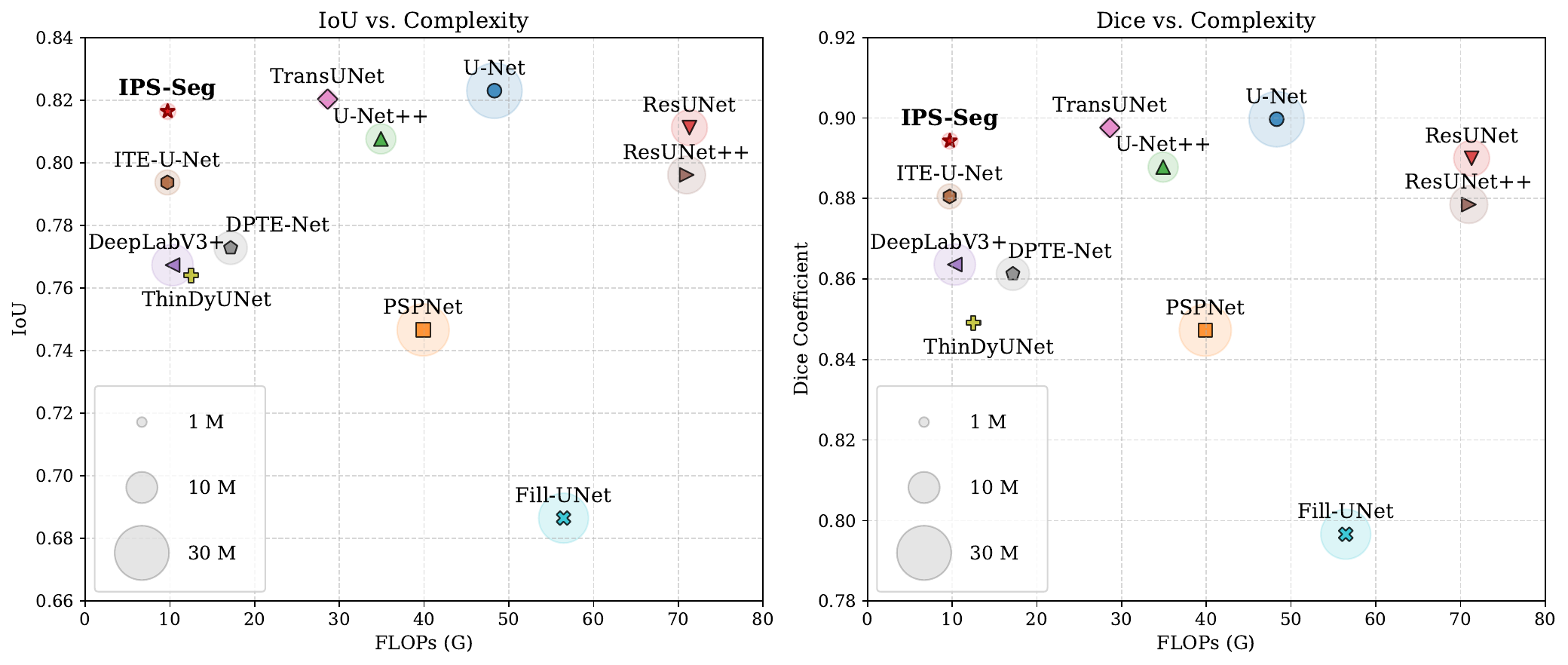}
\caption{Performance-efficiency trade-off of various segmentation networks on the UAV Semantic Segmentation dataset~\cite{kim2025semantic}.}
\label{fig:tradeoff}
\end{figure}

\begin{itemize}

\item A SAM3-guided pseudo-label supervision framework is proposed for UAV target segmentation under annotation-scarce conditions. The framework investigates both direct pseudo-label generation and an iterative refinement strategy that combines open-vocabulary localization with patch-wise segmentation to improve output mask quality.

\item A lightweight segmentation network, termed IPS-Seg, is introduced by combining an IdentityFormer backbone, an ASPP bottleneck, and a PixelShuffle decoder to achieve efficient and accurate UAV target segmentation.

\item Experimental results demonstrate that SAM3-generated pseudo-labels provide effective supervision for lightweight segmentation networks, and the proposed IPS-Seg achieves a favorable balance between segmentation accuracy and computational efficiency.

\end{itemize}

The remainder of this paper is organized as follows. Section~\ref{sec:related} reviews related work. Section~\ref{sec:method} presents the proposed methodology. Section~\ref{sec:experiments} reports the experimental results and ablation studies. Section~\ref{sec:conclusion} concludes the paper.

\section{Related Work}
\label{sec:related}

\subsection{UAV Target Segmentation}

Encoder-decoder architectures have become the dominant framework for dense prediction tasks in computer vision \cite{ronneberger2015u,tran2019robust,tran2022novel,tran2025encoder}, owing to their ability to capture high-level semantic representations while progressively recovering spatial details. Semantic segmentation, as one of the most important dense prediction tasks, has benefited significantly from this paradigm through architectures such as U-Net~\cite{ronneberger2015u} and DeepLab~\cite{chen2017deeplab}. Despite their success, UAV target segmentation remains particularly challenging due to the small object size, rapid motion, and significant appearance variations caused by changes in viewpoints and environmental conditions. To address these challenges, recent studies have explored multi-scale feature aggregation and contextual modeling strategies. Architectures based on SPP~\cite{he2015spatial} and ASPP~\cite{chen2017deeplab} have been widely adopted to enhance feature representations across multiple scales, while transformer-based models \cite{chen2021transunet} further improve global context modeling through self-attention mechanisms. Despite these advances, achieving robust segmentation performance in UAV scenarios often requires large amounts of annotated data, motivating research on learning under limited supervision.

\subsection{Segmentation under Limited Supervision}

The scarcity of pixel-level annotations in UAV imagery has motivated extensive research on learning paradigms that reduce the reliance on dense supervision. Among them, semi-supervised and self-supervised frameworks have emerged as two prominent directions. Semi-supervised methods leverage a small set of labeled samples together with abundant unlabeled data. Representative approaches include consistency regularization~\cite{french2019semi}, cross-consistency training~\cite{ouali2020semi}, and pseudo-labeling~\cite{lee2013pseudo}, where predictions generated from unlabeled data are utilized as additional supervisory signals to improve model generalization. On the other hand, self-supervised learning aims to learn transferable visual representations directly from unlabeled data. Recent advances have significantly improved representation learning quality and reduced the dependence on manual annotations, with various approaches proposed for semantic segmentation~\cite{araslanov2021self,ziegler2022self}.

\subsection{Foundation Models for UAV Segmentation}

Recent advances in vision foundation models have significantly improved the transferability of visual representations across diverse downstream tasks. Notably, the Segment Anything Model (SAM)~\cite{kirillov2023segment} introduced a promptable segmentation framework capable of generating high-quality object masks with impressive zero-shot generalization. Subsequent versions, including SAM2~\cite{ravi2025sam} and SAM3~\cite{carion2025sam}, have further enhanced segmentation quality, prompt flexibility, and open-vocabulary understanding, enabling robust object localization without the need for task-specific retraining. However, directly applying SAMs to UAV imagery remains challenging due to the small object size, complex backgrounds, and significant appearance variations commonly encountered in aerial scenes, often resulting in inaccurate or fragmented masks. Furthermore, the high computational complexity of these foundation models limits their deployment on resource-constrained UAV platforms. These limitations have shifted the research focus toward leveraging foundation models as powerful annotation sources. Recent studies have successfully utilized SAM-generated masks for annotation assistance \cite{wang2023sammed}, pseudo-label generation \cite{jiang2023segment}, and weakly supervised learning \cite{chen2023segment}. Building on this, this work is motivated by the potential of using these models to provide high-quality supervision for training lightweight, task-specific segmentation networks.

\subsection{Research Gap}

Although recent studies have partially reduced the dependence on manual annotations, two challenges remain. First, pseudo-labels generated by foundation models often exhibit inaccurate boundaries and fragmented masks when applied to small and distant objects like UAV targets. Second, many high-performance segmentation models remain computationally demanding for deployment on resource-constrained UAV platforms. To address these challenges, a unified framework is introduced that leverages SAM3-guided pseudo-label generation to provide effective supervision for training a lightweight UAV target segmentation model which can achieve a favorable balance between segmentation accuracy and computational efficiency.

\begin{figure}
\centering
\includegraphics[width=\textwidth]{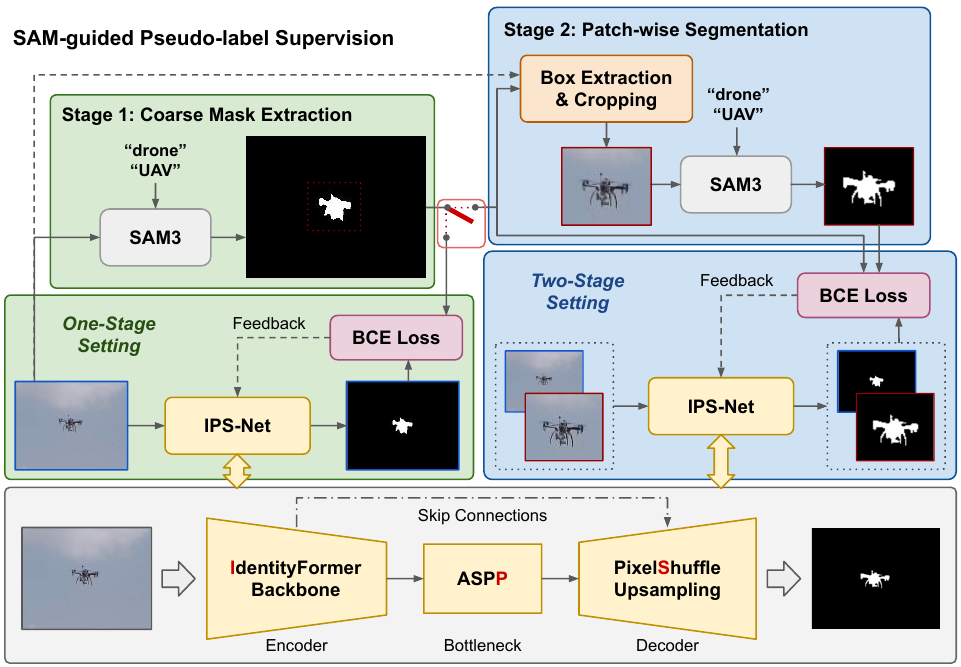}
\caption{The proposed framework.}
\label{fig:framework}
\end{figure}

\section{Methodology}
\label{sec:method}

The proposed framework consists of two components: a SAM3-guided pseudo-label generation pipeline and a lightweight segmentation network, termed \textbf{IPS-Seg}. Two supervision strategies are considered to investigate the effect of pseudo-label quality. The first directly adopts unaltered masks predicted by SAM3 as pseudo-labels, whereas the second localizes candidate UAV regions and re-applies SAM3 on object-centric image patches to obtain finer masks. Since UAVs typically occupy only a small portion of the image, the patch-based strategy enables more accurate boundary delineation and better preservation of fine structural details. The resulting pseudo-labels are then used to optimize IPS-Seg, which adopts an IdentityFormer backbone for lightweight feature extraction, an ASPP bottleneck for multi-scale context aggregation, and a PixelShuffle decoder for progressive spatial resolution recovery. An overview of the framework is illustrated in Fig.~\ref{fig:framework}.

\begin{figure}
\centering
\includegraphics[width=\textwidth]{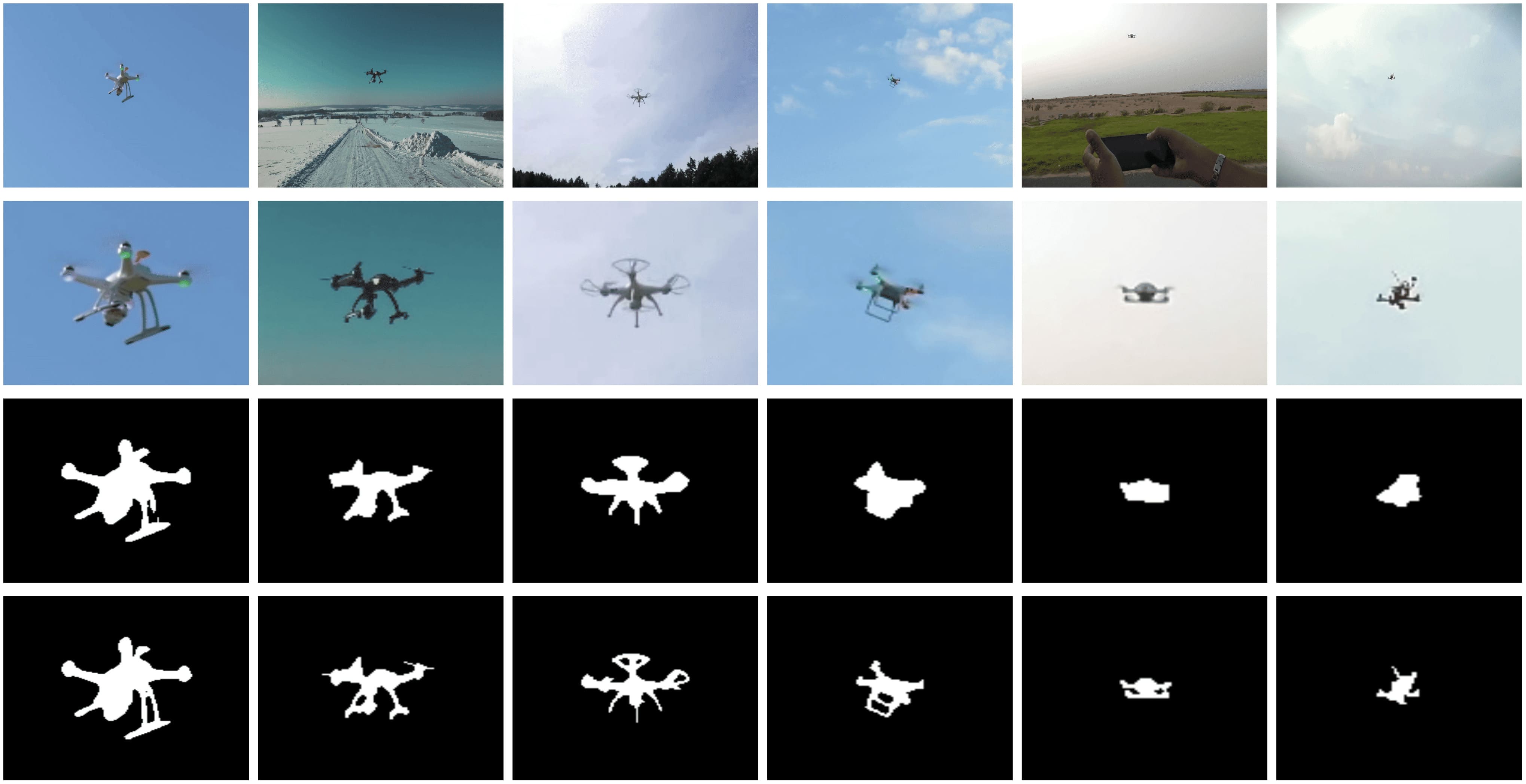}
\caption{Visual comparison of masks generated by SAM3 and the proposed two-stage strategy. From top to bottom: full-size input image, target region, direct prediction by SAM3 (zoomed-in), and fine mask produced by the two-stage approach.}
\label{fig:2stage_labeling}
\end{figure}

\subsection{Framework Overview}

Let the unlabeled training data set be defined as:
\begin{equation}
\mathcal{D}=\{x_j\}_{j=1}^{N},
\label{eq:unlabeled_dataset}
\end{equation}
where $x_j$ denotes an input RGB image. For each image, a pseudo-label $\hat{y}_j$ is generated by:
\begin{equation}
\hat{y}_j=\Phi_{\mathrm{SAM3}}(x_j),
\label{eq:pseudo_label_gen}
\end{equation}
where $\Phi_{\mathrm{SAM3}}(\cdot)$ denotes the SAM3-guided pseudo-label generation process. Depending on the supervision strategy, $\Phi_{\mathrm{SAM3}}(\cdot)$ corresponds to either direct SAM3 prediction or the proposed two-stage generation pipeline. The generated pseudo-labels are paired with the corresponding input images to construct the training data set:
\begin{equation}
\hat{\mathcal{D}}
=
\{(x_j,\hat{y}_j)\}_{j=1}^{N},
\label{eq:training_set}
\end{equation}
which is used to optimize the lightweight segmentation network IPS-Seg:
\begin{equation}
f_{\theta}:x\rightarrow \hat{y},
\label{eq:seg_model}
\end{equation}
where $\theta$ denotes the network parameters. The pseudo-label generation procedure and the IPS-Seg architecture are described in the following subsections.

\begin{figure}
\centering
\includegraphics[width=\textwidth]{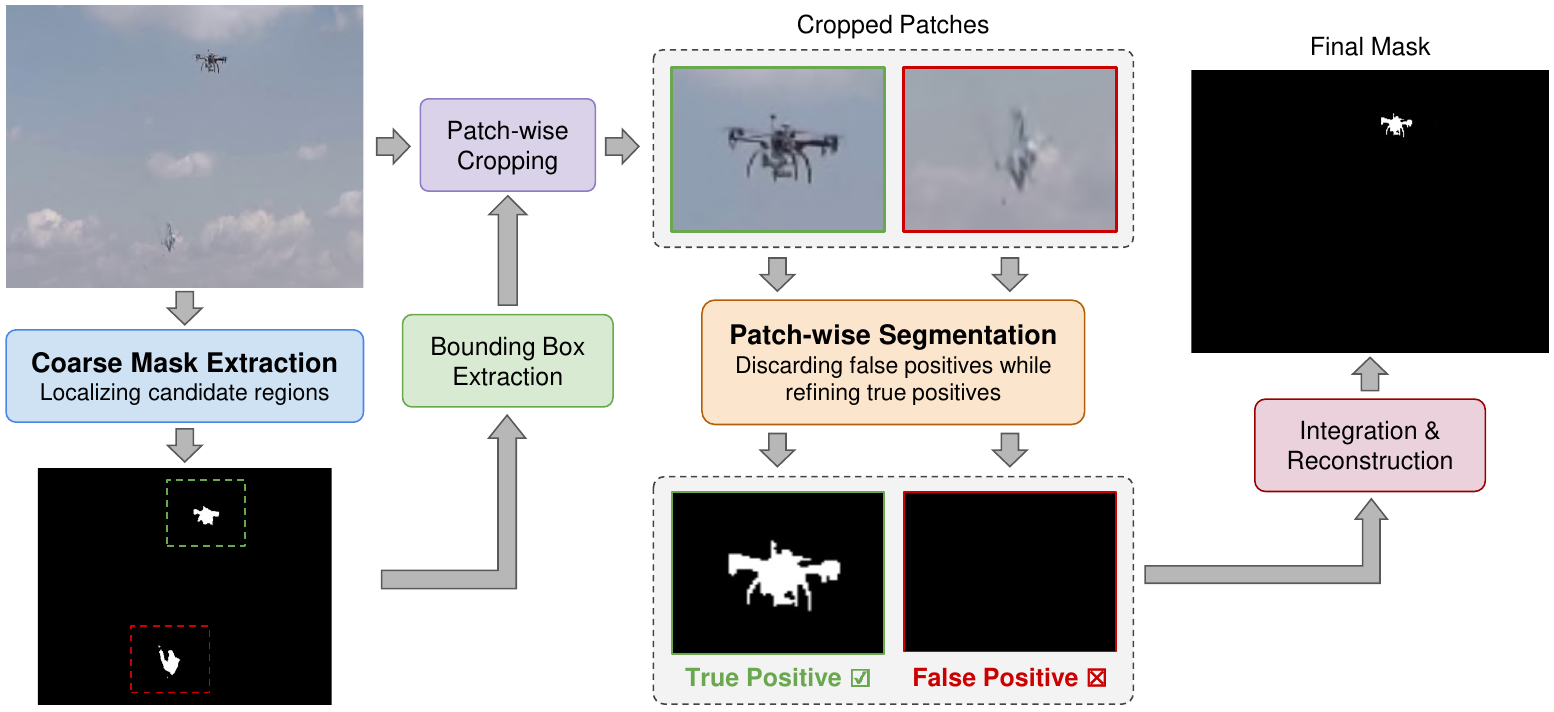}
\caption{Illustration of the proposed two-stage generation/prediction strategy. Candidate regions extracted from coarse masks are re-segmented to discard false positives and improve object boundaries.}
\label{fig:fp_filtering}
\end{figure}

\subsection{SAM3-Guided Pseudo-Label Generation}

Given an unlabeled image $x_j \in \mathcal{D}$, pseudo-labels are generated using either a one-stage or a two-stage SAM3-guided strategy.

\subsubsection{One-Stage Strategy}

The one-stage strategy directly applies SAM3 to the input image:
\begin{equation}
\hat{y}_j
=
m_j^{\mathrm{coarse}}
=
\mathrm{SAM3}(x_j,\mathcal{P}),
\label{eq:one_stage}
\end{equation}
where $\mathcal{P}$ denotes a set of text prompts describing UAV-related concepts (e.g., ``drone'', ``UAV'', and ``quadcopter''). The predicted coarse mask $m_j^{\mathrm{coarse}}$ is directly adopted as the pseudo-label for optimizing IPS-Seg in this training setting.

\subsubsection{Two-Stage Strategy}

To improve pseudo-label quality, a refinement stage is introduced after the initial SAM3 prediction. Given the coarse mask $m_j^{\mathrm{coarse}}$, the target candidate regions are first localized by:
\begin{equation}
C_j=\mathcal{B}(m_j^{\mathrm{coarse}}),
\end{equation}
where $\mathcal{B}(\cdot)$ denotes the bounding-box extraction and cropping operator with a small spatial padding to preserve surrounding contextual information, and $C_j$ represents the set of candidate object-centric cropped patches. These localized patches are subsequently re-segmented by SAM3:
\begin{equation}
M_j^{\mathrm{crop}}
=
\mathrm{SAM3}(C_j,\mathcal{P}),
\end{equation}
where $M_j^{\mathrm{crop}}$ denotes the set of patch-wise segmentation masks with finer object contours. As illustrated in Fig. \ref{fig:2stage_labeling}, the SAM3 baseline (third row) often yields coarse results that miss fine structural details when applied to full-size images, in contrast, the proposed approach (fourth row) effectively captures delicate components like propellers and landing gears.

Unlike previous SAM models, SAM3 incorporates a \textit{presence head} that predicts whether a queried object exists in the input image. This capability enables each cropped candidate region to be verified, allowing false detections produced during the coarse stage to be discarded while preserving valid object regions, as illustrated in Fig.~\ref{fig:fp_filtering}. Such verification is particularly beneficial for small UAVs and background objects with similar visual characteristics. As described in Fig. \ref{fig:framework}, during the two-stage strategy, the original image $x_j$ is supervised by the coarse mask $m_j^{\mathrm{coarse}}$, whereas the cropped object-centric patches $C_j$ are supervised by the patch-wise masks $M_j^{\mathrm{crop}}$. Consequently, IPS-Seg simultaneously learns global object localization from coarse supervision as well as the capability of patch-wise verification and fine-grained boundary delineation from patch-level supervision.

During inference in the two-stage setting, IPS-Seg follows the same hierarchical prediction strategy. The network first predicts a coarse segmentation mask from the full image to localize candidate regions. The cropped regions are then processed again by IPS-Seg to generate refined segmentation masks while verifying object existence. The refined predictions are finally projected back to the original image coordinates to obtain the final segmentation result, as illustrated in Fig.~\ref{fig:fp_filtering}.

\subsection{IPS-Seg}
\label{subsec:ipsseg}

The overall architecture of IPS-Seg follows an $encoder\rightarrow bottleneck\rightarrow decoder$ structure with a four-stage design to better preserve spatial information for small targets while reducing computational complexity. The network name reflects its core components: an \textbf{Identity}Former backbone \cite{yu2023metaformer} for feature extraction, Atrous Spatial Pyramid \textbf{Pooling} (ASPP) \cite{chen2017deeplab} for multi-scale context aggregation, and Pixel\textbf{Shuffle}-based decoding \cite{shi2016real} for progressive spatial resolution recovery, as illustrated in Fig. \ref{fig:framework}. Skip connections are incorporated between encoder and decoder stages to preserve fine-grained spatial information. 

\subsubsection{IdentityFormer Backbone}
The encoder adopts an IdentityFormer backbone based on the MetaFormer paradigm \cite{yu2023metaformer}. Unlike conventional transformer-based models that rely on computationally intensive self-attention \cite{chen2021transunet,dosovitskiy2020image}, IdentityFormer employs an identity token mixer to achieve efficient feature extraction with minimal computational overhead. Given an input feature map $X$, the token-mixing operation is defined as:
\begin{equation}
\mathrm{TokenMixer}(X) = \mathrm{IdentityMapping}(X) = X,
\label{eq:token_mixer}
\end{equation}
where the identity mapping preserves feature representations without introducing additional spatial mixing operations.
Despite its simplicity, IdentityFormer maintains strong representation capability through residual connections, normalization layers, and channel-wise feature transformations, making it suitable for small target segmentation tasks \cite{do2025improved}.

\subsubsection{ASPP Bottleneck}
To enhance contextual representation, the bottleneck incorporates ASPP \cite{chen2017deeplab} for multi-scale feature aggregation. The ASPP module applies parallel atrous convolutions with different dilation rates:
\begin{equation}
\mathrm{ASPP}(X) = \mathrm{Concat} \left( A_1(X), A_2(X), \ldots, A_k(X) \right),
\label{eq:aspp}
\end{equation}
where $A_k(\cdot)$ denotes an atrous convolution, and $\mathrm{Concat}(\cdot)$ represents channel-wise concatenation.
By aggregating features from multiple receptive fields, ASPP improves robustness to scale variations commonly encountered in UAV imagery.

\subsubsection{PixelShuffle Decoder}
The decoder progressively restores spatial resolution using PixelShuffle-based upsampling \cite{shi2016real}:
\begin{equation}
X_{\mathrm{up}} = \mathrm{PixelShuffle}(X).
\label{eq:pixel_shuffle}
\end{equation}
Skip connections between corresponding encoder-decoder levels are incorporated to preserve fine-grained spatial details and improve boundary reconstruction. Compared with transposed convolutions, PixelShuffle provides computationally efficient upsampling while reducing checkerboard artifacts.

\subsection{Training Objective}
The proposed IPS-Seg network is trained using pseudo-labels obtained from the unlabeled dataset. The network prediction is represented as:
\begin{equation}
\tilde{y}_j = f_{\theta}(x_j),
\label{eq:prediction}
\end{equation}
where $f_{\theta}(\cdot)$ denotes IPS-Seg parameterized by $\theta$.
The optimization process is performed using Binary Cross-Entropy (BCE) loss between the predicted mask $\tilde{y}_j$ and the pseudo-label $\hat{y}_j$:
\begin{equation}
\mathcal{L} = \mathcal{L}_{\mathrm{BCE}} \left( \tilde{y}_j, \hat{y}_j \right).
\label{eq:bce_loss}
\end{equation}
Through pseudo-label supervision, the framework transfers segmentation knowledge from SAM3 into a lightweight architecture suitable for efficient deployment.

\section{Experiments}
\label{sec:experiments}

\subsection{Experimental Setup}

This section describes the experimental protocols used to evaluate the proposed SAM3-guided pseudo-label supervision framework and the IPS-Seg architecture under both fully supervised and pseudo-supervised learning settings. The fully supervised experiments assess the intrinsic capability of the proposed IPS-Seg architecture using annotated masks, whereas the pseudo-supervised experiments evaluate the effectiveness of the proposed SAM3-guided supervision framework under annotation-scarce conditions.

\begin{figure}
\centering
\includegraphics[width=\textwidth]{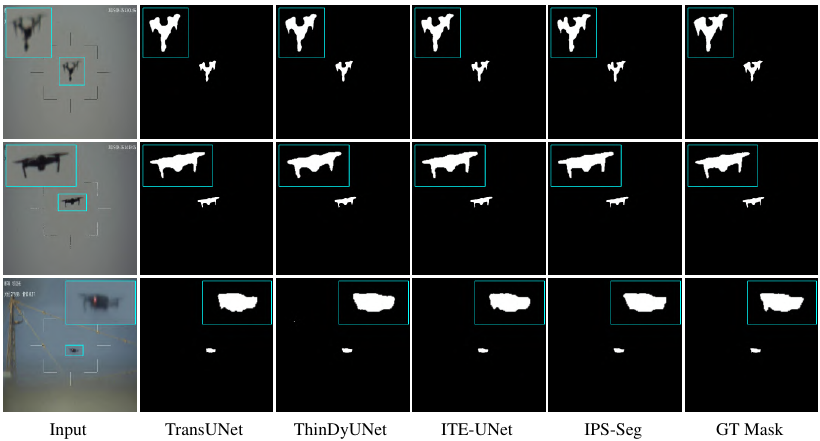}
\caption{Qualitative comparison of different segmentation approaches.}
\label{fig:qualitative}
\end{figure}

\begin{table}[t]
\centering
\caption{Quantitative performance comparison with state-of-the-art methods.}
\label{tab:segmentation_results}
\begin{tabular}{cccccc}
\hline
\multirow{2}{*}{Model} & \multirow{2}{*}{Type} & \multicolumn{2}{c}{Accuracy} & \multicolumn{2}{c}{Complexity} \\
\cline{3-6}
 &  & IoU & Dice & Params (M) & FLOPs (G) \\
\hline
U-Net \cite{ronneberger2015u}       & supervised & 0.8230 & 0.8997 & 31.04 & 48.30 \\
PSPNet \cite{zhao2017pyramid}       & supervised & 0.7466 & 0.8473 & 27.50 & 39.88 \\
U-Net++ \cite{zhou2018unet++}       & supervised & 0.8076 & 0.8878 & 9.16  & 34.90 \\
ResUNet \cite{zhang2018road}        & supervised & 0.8112 & 0.8900 & 13.04 & 71.32 \\
DeepLabV3+ \cite{chen2018encoder}   & supervised & 0.7673 & 0.8636 & 17.42 & 10.30 \\
ResUNet++ \cite{jha2019resunet++}   & supervised & 0.7961 & 0.8785 & 14.48 & 70.99 \\
TransUNet \cite{chen2021transunet}  & supervised & 0.8203 & 0.8976 & 3.63  & 28.61 \\
DPTE-Net \cite{tran2025distilled}   & supervised & 0.7728 & 0.8613 & 11.14 & 17.16 \\
ThinDyUNet \cite{kim2025semantic}   & supervised & 0.7667 & 0.8491 & 0.81  & 12.49 \\
Fill-UNet \cite{liu2025fill}        & supervised & 0.6865 & 0.7966 & 25.24 & 56.46 \\
ITE-U-Net \cite{do2025improved}     & supervised & 0.7937 & 0.8805 & 6.27 & 9.69 \\
\textbf{IPS-Seg}                    & supervised & 0.8164 & 0.8943 & 2.69 & 9.72 \\
\hline
\end{tabular}
\end{table}

\textbf{Dataset:} Experiments are conducted on the UAV Semantic Segmentation dataset~\cite{kim2025semantic}, which contains over 300K RGB aerial images with pixel-level semantic annotations collected under diverse viewpoints, flight altitudes, illumination conditions, object scales, and background complexities. To reduce computational cost while maintaining representative evaluation coverage, approximately 8K images are randomly sampled from the dataset and divided into training and validation sets with a 90\%/10\% split. Two training settings are considered. In the \emph{fully supervised} setting, the training images are optimized using the corresponding ground-truth segmentation masks. In the \emph{pseudo-supervised} setting, all ground-truth annotations of the training set are discarded, and SAM3-generated pseudo-labels are used as supervisory signals instead. The validation set always retains the original annotated labels and is used exclusively for evaluation.

\textbf{Implementation Details:}
The proposed IPS-Seg is implemented in PyTorch and trained using the Adam optimizer with an initial learning rate of $1\times10^{-4}$. A cosine annealing schedule is adopted to gradually decay the learning rate during optimization. All models are trained for 100 epochs using input images resized to $256\times256$. Experiments are conducted on a single NVIDIA RTX 3080 GPU.

\textbf{Evaluation Metrics:}
Segmentation accuracy is evaluated using the Intersection over Union (IoU) and Dice coefficient. To assess computational efficiency, the number of trainable parameters (Params) and floating-point operations (FLOPs) are also reported. Together, these metrics provide a comprehensive evaluation of the trade-off between segmentation performance and computational complexity.

\subsection{Results and Analysis}

\subsubsection{Supervised Learning Setting}

The fully supervised setting provides a baseline evaluation of the proposed IPS-Seg architecture independent of the pseudo-label generation framework. Quantitative comparisons with representative CNN-, transformer-, and hybrid-based segmentation methods are summarized in Table~\ref{tab:segmentation_results}.

IPS-Seg achieves an IoU of 0.8164 and a Dice score of 0.8943 while requiring only 2.69M parameters and 9.72 GFLOPs. Although U-Net attains a slightly higher IoU (0.8230), it contains more than $11\times$ parameters (31.04M) and nearly $5\times$ computational cost (48.30 GFLOPs). Similar observations can be made for ResUNet, ResUNet++, and U-Net++, whose competitive segmentation accuracy is achieved at substantially higher computational complexity. Compared with other lightweight models, IPS-Seg consistently delivers superior performance. Specifically, IPS-Seg improves the IoU by 4.97\% and 2.27\% over ThinDyUNet and ITE-U-Net, respectively, while maintaining a comparable model size and computational cost. Compared with TransUNet, IPS-Seg achieves competitive accuracy using approximately 26\% fewer parameters and about one-third of the computational complexity. These results demonstrate that the proposed architectural combination offers an effective balance between segmentation accuracy and computational efficiency, as previously illustrated in Fig. \ref{fig:tradeoff}.

Qualitative comparisons are also presented in Fig.~\ref{fig:qualitative}. Overall, all methods successfully identify the UAV regions; however, noticeable differences can be observed in boundary quality and prediction consistency. The proposed IPS-Seg produces segmentation masks that closely match the ground-truth annotations across different scales and imaging conditions. In contrast, ThinDyUNet occasionally introduces small false-positive regions, as illustrated in the third example, indicating that its feature representations are more susceptible to background interference. ITE-U-Net and TransUNet generally produce reasonable masks but exhibit slightly less accurate object boundaries, particularly for small or blurred UAV targets. These visual results are consistent with the quantitative evaluation and demonstrate that IPS-Seg achieves more reliable segmentation with fewer spurious predictions.

\begin{figure}
\centering
\includegraphics[width=\textwidth]{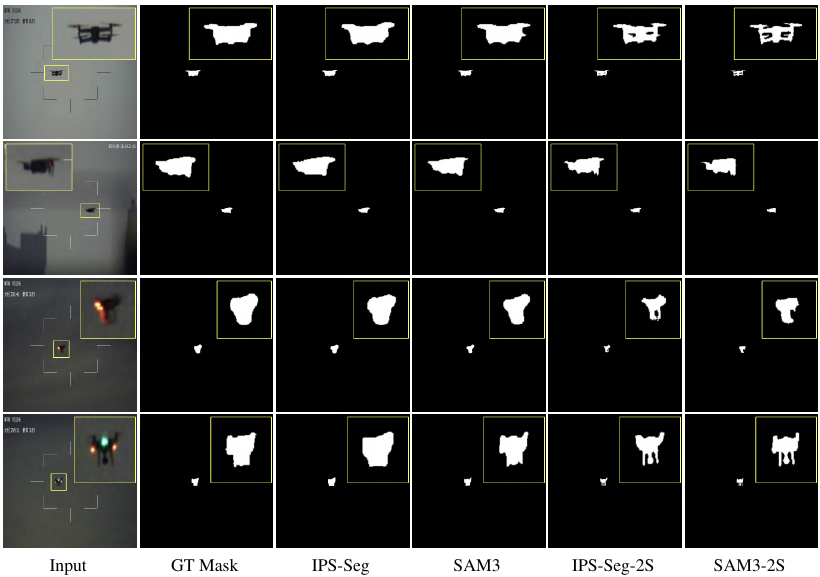}
\caption{Qualitative comparison between the one-stage and two-stage strategies.}
\label{fig:mask_refinement_results}
\end{figure}

\begin{table}[t]
\centering
\caption{Quantitative performance comparison in the one-stage and two-stage strategies.}
\label{tab:segmentation_results_2}
\begin{tabular}{ccccc}
\hline
\multirow{2}{*}{Model} & \multirow{2}{*}{Teacher} & \multirow{2}{*}{Supervision} & \multicolumn{2}{c}{Accuracy}  \\
\cline{4-5}
 & & & IoU & Dice \\
\hline
SAM3 & - & zero-shot & 0.8201 & 0.8991 \\
\textbf{IPS-Seg} & SAM3  & pseudo-labeling & 0.7941 & 0.8737 \\
\hline
SAM3-2S & - & zero-shot & 0.6656 & 0.7945 \\
\textbf{IPS-Seg-2S} & SAM3-2S & pseudo-labeling & 0.6002 & 0.7291 \\
\hline
\end{tabular}
\end{table}

\subsubsection{SAM3-Guided Learning Setting}

This section evaluates the performance of IPS-Seg when it is trained exclusively using pseudo-labels generated by the SAM3-guided supervision framework. Quantitative comparisons are reported in Table~\ref{tab:segmentation_results_2}, while representative qualitative results are presented in Fig.~\ref{fig:mask_refinement_results}.

As shown in Table~\ref{tab:segmentation_results_2}, the zero-shot SAM3 model achieves an IoU of 0.8201 and a Dice score of 0.8991, demonstrating the strong generalization capability of SAM3 without task-specific fine-tuning. In this configuration, IPS-Seg achieves an IoU of 0.7941 and a Dice score of 0.8737, retaining most of the teacher's segmentation performance despite its substantially lower computational complexity. This result indicates that the proposed supervision framework effectively transfers the segmentation knowledge of SAM3 to a compact network suitable for deployment on resource-constrained platforms.

The effectiveness of the proposed two-stage strategy is also investigated. Quantitatively, the two-stage SAM3 approach (SAM3-2S) obtains lower overlap scores than direct SAM3 prediction, achieving an IoU of 0.6656 and a Dice score of 0.7945, while the corresponding two-stage student model (IPS-Seg-2S) achieves an IoU of 0.6002 and a Dice score of 0.7291. At first glance, these results appear inferior to those obtained using direct pseudo-label supervision. However, a different conclusion can be drawn from the qualitative comparisons in Fig.~\ref{fig:mask_refinement_results}. As shown in Fig.~\ref{fig:mask_refinement_results}, the one-stage masks generated by SAM3 and IPS-Seg generally preserve the coarse object shape but tend to produce relatively smooth boundaries that omit fine structural details. In contrast, the two-stage strategy recovers much richer object structures such as the thin landing gears and the empty regions between UAV components. These refined masks more faithfully represent the actual UAV geometry, especially for complex quadcopter structures. Similar improvements are consistently observed for both the teacher (SAM3-2S) and the student (IPS-Seg-2S), indicating that the refined pseudo-labels successfully transfer fine-grained structural information during training.

The apparent discrepancy between the quantitative and qualitative results can largely be attributed to the characteristics of the dataset annotations. The ground-truth masks primarily provide coarse object boundaries, whereas the refined predictions preserve finer structural details that are not explicitly annotated. Since IoU and Dice evaluate pixel-wise overlap with the provided annotations, these additional structural details may be penalized despite producing visually more faithful segmentation outcomes. This limitation is particularly pronounced for small UAV targets, where slight boundary variations correspond to a relatively large proportion of the object area. Consequently, conventional overlap-based metrics may underestimate the quality of fine-grained segmentation produced by the proposed refinement strategy, suggesting that evaluation protocols for fine-grained target segmentation may benefit from boundary-aware metrics in addition to conventional region-overlap measures.

\begin{table}[t]
\centering
\caption{Ablation study on overall architecture.}
\label{tab:ablation_main}
\begin{tabular}{ccccccc}
\hline
\multicolumn{3}{c}{Architecture} & \multicolumn{2}{c}{Accuracy} & \multicolumn{2}{c}{Complexity} \\
\cline{1-7}
 Backbone & Bottleneck & Decoder & IoU & Dice & Params (M) & FLOPs (G) \\
\hline
Conv & - & UpSample & 0.7165 & 0.8035 & 1.60 & 7.67 \\
Conv & SPP & UpSample & 0.7335 & 0.8274 & 1.65 & 7.73 \\
IndentityFormer & SPP & UpSample & 0.7873 & 0.8740 & 1.56 & 7.43 \\
IndentityFormer & ASPP & UpSample & 0.7997 & 0.8802 & 1.82 & 7.69 \\
IndentityFormer & ASPP & PixelShuffle & \textbf{0.8164} & \textbf{0.8943} & 2.69 & 9.72 \\
\hline
\end{tabular}
\end{table}

\subsection{Ablation Study}
\label{subsec:ablation}

To further analyze the contribution of each component in the proposed IPS-Seg architecture, comprehensive ablation studies are conducted on both the overall network design and the encoder backbone under the fully supervised setting using identical training protocols.

\subsubsection{Overall Network Architecture}

Table~\ref{tab:ablation_main} summarizes the incremental development of IPS-Seg, starting from a four-stage U-Net baseline without a contextual aggregation bottleneck. From the baseline, introducing the SPP bottleneck \cite{he2015spatial} improves the IoU from 0.7165 to 0.7335 and the Dice score from 0.8035 to 0.8274 with marginal additional computational cost. This improvement indicates that enlarging the receptive field through multi-scale feature aggregation is beneficial when object scales vary considerably across different scenes.

Replacing the convolutional encoder with the IdentityFormer backbone improves the IoU to 0.7873 and the Dice score to 0.8740 while slightly reducing computational complexity. This suggests that the MetaFormer-based feature representation is considerably more effective than conventional convolutional feature extraction for capturing discriminative object characteristics in this task.

Replacing SPP with ASPP further improves the IoU to 0.7997 and the Dice score to 0.8802, demonstrating that parallel atrous convolutions provide richer multi-scale contextual representations than conventional spatial pooling. Such contextual modeling is particularly advantageous for UAV imagery, where the target size changes significantly due to varying camera viewpoints and imaging distances.

Finally, replacing UpSampling with PixelShuffle in the decoder yields the best overall performance, achieving an IoU of 0.8164 and a Dice score of 0.8943. Although the model complexity increases modestly to 2.69M parameters and 9.72 GFLOPs, the additional computational overhead is accompanied by a noticeable improvement in segmentation accuracy. This result indicates that the progressive sub-pixel reconstruction strategy is more effective at recovering fine spatial details.

\begin{table}[t]
\centering
\caption{Ablation study on the encoding backbone.}
\label{tab:ablation_encoder}
\begin{tabular}{ccccc}
\hline
\multirow{2}{*}{Backbone} & \multicolumn{2}{c}{Accuracy} & \multicolumn{2}{c}{Complexity} \\
\cline{2-5}
 & IoU & Dice & Params (M) & FLOPs (G) \\
\hline
Residual & 0.6819 & 0.7917 & 3.67 & 12.79 \\
PoolFormer & 0.7041 & 0.8089 & 2.69 & 9.73 \\
ConvFormer & 0.7494 & 0.8452 & 2.71 & 9.90 \\
\textbf{IndentityFormer} & \textbf{0.8164} & \textbf{0.8943} & 2.69 & 9.72 \\
\hline
IndentityFormer + SCConv & 0.7803 & 0.8692 & 2.97 & 9.86 \\
IndentityFormer + CAFormer & 0.7828 & 0.8705 & 2.69 & 9.74 \\
IndentityFormer + DualKAN & 0.7437 & 0.8357 & 6.39 & 8.97 \\
\hline
\end{tabular}
\end{table}

\subsubsection{Encoding Backbone}

To further investigate the influence of feature encoding on segmentation performance, different backbone variants are evaluated while keeping the ASPP bottleneck and PixelShuffle decoder unchanged. The quantitative results are summarized in Table~\ref{tab:ablation_encoder}.

The first group compares several representative backbone architectures. The conventional residual encoder achieves an IoU of 0.6819 and a Dice score of 0.7917. Replacing the residual blocks with MetaFormer-based architectures consistently improves segmentation performance. PoolFormer increases the IoU to 0.7041 and the Dice score to 0.8089, while ConvFormer further improves the performance to 0.7494 IoU and 0.8452 Dice. Among all candidates, the IdentityFormer achieves the best performance, obtaining an IoU of 0.8164 and a Dice score of 0.8943 with only 2.69M parameters and 9.72 GFLOPs. Compared with ConvFormer, IdentityFormer improves the IoU while exhibiting slightly lower computational complexity, demonstrating that IdentityFormer provides a more effective feature representation for UAV segmentation than the other variants.

To examine whether more sophisticated feature extraction approaches can further enhance the encoder, three representative modules, including DualKAN~\cite{tran2026unpaired}, SCConv~\cite{liu2020improving}, and CAFormer~\cite{yu2023metaformer} are incorporated into the deeper stages of backbone (stages 3 and 4) where the feature maps become semantically richer. Contrary to expectation, none of these modules improve segmentation performance. IdentityFormer$^{1,2}$+DualKAN$^{3,4}$ increases the model size substantially from 2.69M to 6.39M parameters while reducing the IoU to 0.7437. Likewise, IdentityFormer$^{1,2}$+SCConv$^{3,4}$ and IdentityFormer$^{1,2}$+CAFormer$^{3,4}$ achieve IoU scores of 0.7803 and 0.7828, respectively, both falling below the full IdentityFormer configuration.

These observations suggest that increasing architectural complexity does not necessarily result in better segmentation performance for UAV imagery. Since UAV targets typically occupy only a small fraction of the image, discriminative feature extraction appears to rely more on preserving fine spatial information than on increasing feature transformation capacity. Excessively complex modules may introduce redundant representations, optimization difficulty, or overfitting without providing additional discriminative cues. In contrast, the lightweight IdentityFormer backbone offers a favorable balance between representation capability and computational efficiency, allowing subsequent modules to effectively exploit the extracted features.

\section{Conclusion}
\label{sec:conclusion}

This paper presents a SAM3-guided pseudo-label supervision framework for UAV target segmentation under annotation-scarce conditions. By leveraging SAM3-generated pseudo-labels, the proposed framework enables the training of IPS-Seg, a lightweight segmentation network that integrates an IdentityFormer backbone, an ASPP bottleneck, and a PixelShuffle decoder, thereby achieving accurate and efficient UAV target segmentation.
Extensive experiments demonstrate that the lightweight network design achieves an excellent balance between segmentation accuracy and computational efficiency for resource-constrained deployment through the proposed pseudo-label supervision strategy, which effectively transfers the segmentation capability of a large vision foundation model to a compact network architecture. The experimental results also show that conventional overlap-based metrics may not fully capture improvements in fine-grained boundary quality.

Overall, this work demonstrates the potential of using vision foundation models as automatic annotation sources for training efficient segmentation networks with limited manual supervision. Future work will investigate confidence-aware pseudo-label selection, boundary-aware optimization, and more robust refinement strategies to further improve pseudo-label quality and performance.

\section*{Source code}
The source code of this work is available at \url{https://github.com/tranleanh/ips-seg}.

\medskip

{
\small
\bibliographystyle{unsrtnat}
\bibliography{ref}
}

\end{document}